\pdfoutput=1

\documentclass[11pt]{article}

\usepackage[final]{acl}

\usepackage{times}
\usepackage{latexsym}

\usepackage[T1]{fontenc}

\usepackage[utf8]{inputenc}

\usepackage{microtype}

\usepackage{inconsolata}

\usepackage{graphicx}

\usepackage{subcaption}             
\usepackage{multirow}               
\usepackage{amsmath}                
\usepackage{wasysym}
\usepackage{changepage,threeparttable} 

\newcommand{\note}[1]{\color{black}#1}
\newcommand{\noteb}[1]{\color{black}#1}
\newcommand{\notec}[1]{\color{black}#1}

\newcommand{\new}[1]{#1}

\title{Simple is not Enough: Document-level Text Simplification using Readability and Coherence}

\author{Laura Vásquez-Rodríguez$^{1,2,\ast}$, Nhung T.H. Nguyen$^{2}$, \\ {\bf Piotr Przybyła}$^{3,6}$, {\bf Matthew Shardlow$^4$}, {\bf Sophia Ananiadou}$^{2,5}$\\
  $^1$Idiap Research Institute, Martigny, Switzerland \\
  $^2$National Centre for Text Mining, Department of Computer Science, \\The University of Manchester, Manchester, UK\\
  $^3$Universitat Pompeu Fabra, Barcelona, Spain \\
  $^4$Department of Computing and Mathematics, Manchester Metropolitan University, UK \\
  $^5$Artificial Intelligence Research Center (AIRC), Tokyo, Japan \\
  $^6$Institute of Computer Science, Polish Academy of Sciences, Warsaw, Poland \\
  \texttt{laura.vasquez@idiap.ch, nhung.nguyen@manchester.ac.uk, m.shardlow@mmu.ac.uk} \\
  \texttt{piotr.przybyla@upf.edu, sophia.ananiadou@manchester.ac.uk} \\
  \\}

\begin{document}
\maketitle
\begin{abstract}
In this paper, we present the \textit{SimDoc} system, a simplification model considering simplicity, readability, and discourse aspects, such as coherence. 
In the past decade, the progress of the Text Simplification (TS) field has been mostly shown at a sentence level, rather than considering paragraphs or documents, a setting from which most TS audiences would benefit. We propose a simplification system that is initially fine-tuned with professionally created corpora. Further, we include multiple objectives during training, considering simplicity, readability, and coherence altogether. Our contributions include the extension of professionally annotated simplification corpora by the association of existing annotations into (complex text, simple text, readability label) triples to benefit from readability during training. Also, we present a comparative analysis in which we evaluate our proposed models in a zero-shot, few-shot, and fine-tuning setting using document-level TS corpora, demonstrating novel methods for simplification. Finally, we show a detailed analysis of outputs, highlighting the difficulties of simplification at a document level.
\def\thefootnote{*}\footnotetext{This work was done as a PhD student at the University of Manchester, United Kingdom.}


\end{abstract}

\section{Introduction}


In recent years, Text Simplification (TS) research has explored methods at a sentence level \citep{Alva_Manchego_2020} without considering the benefits of simplifying paragraphs or documents. Except for \citet{siddharthan-2003-preserving, Stajner_2017}, simplification efforts at a document level have been overlooked, and is not until recent years, that initiatives in this domain started to be developed ~\cite{sun-etal-2021-document, srikanth-li-2021-elaborative, cripwell-etal-2023-document}, together with our document level approach proposed in this paper. Overall, this path is challenging due to data scarcity in all languages and unreliable evaluation metrics. Nevertheless, the community has started to work towards this avenue as well \citep{dmitrieva-tiedemann-2021-creating, rios-etal-2021-new, devaraj-etal-2021-paragraph, stodden-etal-2023-deplain} and to improve previous resources \citep{xu-etal-2015-problems, vajjala-lucic-2018-onestopenglish}. 

We considered three factors for driving simplification models at a document level: simplicity, readability, and coherence. Simplicity is represented by a text that is easy to understand for the target audience \citep{stajner-saggion-2018-data}. In contrast, readability is focused on how legible a document is and we assess this property based on the difficulty of reading a text \citep{vajjala-2022-trends}. Simplification and readability concepts are closely related and often used interchangeably, as a simplified text would also improve its readability while still preserving its meaning \citep{Thanyyan_2021}. Finally, there is coherence \citep{Jurafsky-2021}, which measures logical relationships between sentences, e.g., sentences that follow the same topic rather than \note{being} randomly assembled.

In this work, we develop the \textit{SimDoc} system, combining relevant aspects to simplify text. We enumerate our contributions as follows: 1) An \note{adapted} dataset from professionally annotated corpora (i.e., people with expertise in content simplification) using multiple levels of readability; 2) A comparative analysis using professionally annotated and automatically aligned corpora on T5 model in multiple settings such as zero-shot, few-shot and fine-tuning; 3) A simplification model using a joint measurement of simplification, readability and coherence during training; 4) A detailed analysis of the system outputs, that highlights the limitations and future work of our approach.\footnote{We make our code available in Github: \url{https://github.com/lmvasque/ts-doc}}

\section{Related Work}
\label{sec:related_work}

Aside from the work done in recent years, both models and datasets for TS at a document level have been limited and scarce \citep{alva-manchego-etal-2019-cross}. Document-level simplification inherits the generative nature of closely related tasks such as summarisation and machine translations and hence, its similarity in its methods (e.g., BERT-based).  Starting with the general domain, TS work has developed multiple strategies for English where content is either simplified or extended, considering elaboration as a subtask of simplification. For example, \citet{sun-etal-2021-document} proposed and assessed the D-Wikipedia-based dataset with a vanilla transformer \citep{Vaswani_2017}, BertSumextabs \citep{liu-lapata-2019-text} and BART \citep{lewis-etal-2020-bart}. In a more elaborative setting, \citet{srikanth-li-2021-elaborative} used GPT-2 \citep{radford_language_2019} for content insertion and explanation generation for TS. More recently, \citet{cripwell-etal-2023-document}, developed a system that created a simplification plan to predict simplification operations (e.g., copy, rephrase, split and delete) over sentences in a document.  Work in document-level simplification has also been prominent in the medical domain. \citet{devaraj-etal-2021-paragraph} proposed a paragraph-aligned corpus based on the Cochrane dataset, using BART as a TS system. 

Except for \citet{cripwell-etal-2023-document}, the aforementioned approaches show novel strategies at a document level, but, these do not follow a controllable, multi-task-oriented setting where multiple tasks can be combined during training. The controllability of generative tasks in LLMs has been popular in general \citep{Zhang_2023}, but also in simplification \citep{nishihara-etal-2019-controllable, maddela-etal-2021-controllable, martin-etal-2020-controllable}, and multi-task settings where, in addition, we consider readability and readability combined with summarisation \citep{luo-etal-2022-readability, goldsack-etal-2022-making, guo_retrieval_2024}. These tasks are also considered as a two-step process, when the text is first summarised and further, simplified \citep{Guo_Qiu_Wang_Cohen_2021, shaib-etal-2023-summarizing}. Conversational LLMs have also been key in these domains, proposing novel methods for simplification and readability for specific target levels \citep{farajidizaji-etal-2024-possible}.

It has been proven that the inclusion of linguistic features in readability is relevant \citep{wilkens-etal-2024-exploring}. In the same avenue, we proposed the use of coherence as a novel element that has not been considered before in simplification. This is relevant, as diverse types of TS modifications such as in syntactic simplification can affect the coherence of the models \cite{Abdolahi_2016}. In other domains, coherence has been well explored \cite{lai-tetreault-2018-discourse, mesgar-strube-2018-neural, naismith-etal-2023-automated}, but it has been narrowly applied in simplification research \citep{siddharthan_syntactic_2006, LEROY2013717}. We contribute with our work to this effort through the integration of coherence during training.

Finally, we would like to highlight contributions that are closely related to our work (but still, without the coherence aspect). \citet{sheang-saggion-2021-controllable} considered a text-to-text approach for sentence simplification. Although this effort is different in the sentence-document setting, they validated the use of the T5 model \citep{Colin_2020} for the first time in the task of TS, using control tokens as in \citet{martin-etal-2020-controllable}. We suggest the use of this model for document-level TS with natural language tokens as proposed in the original paper combined with readability and coherence aspects of language. In our work, we will use tags for each task, such as ``simplify'' to perform simplification, rather than adding information about the expected simplification (e.g., using token ``chars 0.5'' to define the length ratio between the source and target sentence) as done in previous work. 

\begin{figure*}[ht!]
\begin{center}
\includegraphics[width=0.70\linewidth]{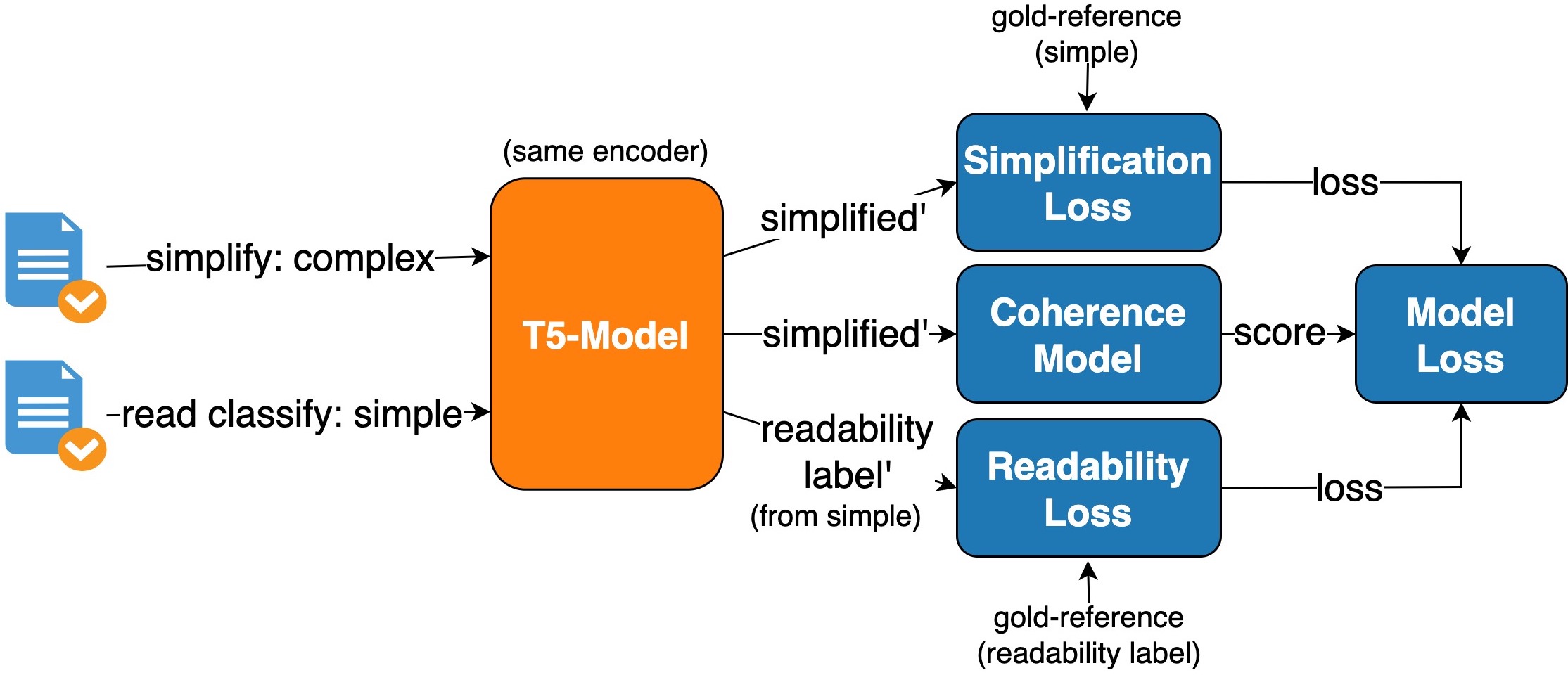}
\caption{TS model architecture. We input the complex (for TS) and the simple gold standard text (for readability) in the model. Predicted simplifications are used in the loss and coherence evaluation. The predicted readability labels are used for readability loss. Finally, we propose a combined loss using TS, readability and coherence.}
\label{fig:combined_loss}
\end{center}
\end{figure*}

\section{Methodology}
\label{methodology}

In this section, we present our methods, starting with the description of our task (Section \ref{task_description}). Further, we present our selected simplification, readability and coherence models for the proposed joint task (Section \ref{sec:models_selection}). Altogether, we merge the capabilities of these models with a customised loss function (Section \ref{sec:loss_function_by_sample}). Finally, we detailed our data selection and preprocessing steps (Section \ref{ch8:data_selection}).


\subsection{Task Description}
\label{task_description}

We proposed the use of the T5 models \citep{Colin_2020} for simplification, which is suitable for controlling different tasks at the same time by the use of control tokens.
\new{These models have an explicit control of the task performed during training. Also, they have been trained in multiple tasks such as summarisation and classification, by explicitly stating the NLP task performed as part of the input.\footnote{Although recent, instruct-based models may seem more proficient for the TS task \citep{wu2024, kew-etal-2023-bless}, they are not suitable for our approach as we focus on architectures that can be deterministic and easily controlled to understand the potential benefit of learning simplification, readability and coherence.}

We leverage the capabilities of text-to-text models with control tokens for simplification generation and readability classification, with an additional coherence reward, supporting a multitasking scenario altogether. }
\note{To achieve this, we appended into the input data the token ``simplify:'' to indicate the simplification task and ``read classify:'' for the readability classification task.} \note{Initially, we experimented with T5-models of different sizes (i.e., T5-small and T5-large) evaluated on a zero, few-shot and fine-tuning scenarios (See Section \ref{appendix:model_settings}), with our customised simplification prompt but with no changes to the T5 model architecture (Section \ref{simple_models})}. 

\note{Further, we introduced a multitasking approach in which we considered simplification and readability altogether (Section \ref{read_models}). Then, we added a coherence reward (Section \ref{coherence_models}) to evaluate the inclusion of coherence into the model training. We trained our models in the aforementioned settings, incrementally evaluating the addition of simplification, readability and coherence into the loss (Section \ref{sec:loss_function_by_sample}). Finally, we also considered additional experiments in which we trained the simplification task as a first step to improve the model performance. See Section \ref{simplification_larger} for more details.}

\subsection{Models}
\label{sec:models_selection}

In this section, we explain our selected models to combine simplification (Section \ref{simple_models}), readability (Section \ref{read_models}) and coherence (Section \ref{coherence_models}) as shown in Figure \ref{fig:combined_loss}.

\subsection{Baseline}
\label{appendix:baseline_models}

\note{We selected the Multilingual Unsupervised Sentence Simplification (MUSS) model \citep{martin-etal-2020-controllable} as the main baseline. For English, the MUSS model was trained in sentence pairs (complex-simple) using either mined paraphrases from the web (\textit{muss\_ en\_mined} model) or from the Wikilarge corpus to perform sentence simplification (\textit{muss\_en\_wikilarge\_mined} model) in a controllable setting. To emulate our document-level setting, we simplified the proposed texts sentence-by-sentence and then, we calculated the evaluation metrics for the whole text by joining all the simplified sentences together. Unlike our model, in which we have to truncate input sentences, MUSS does not require such truncation because each sentence is evaluated individually. Therefore, the FGKL and the FRE scores will slightly vary for the complex text in the baselines in comparison to the proposed systems at a document level.}

\subsubsection{Simplification-based TS Model}
\label{simple_models}

We fine-tuned the proposed T5-models using simplification data (Section \ref{ch8:data_selection}). In this scenario, we evaluated and trained our models in zero, few-shot and fine-tuning scenarios, with no changes to the cross-entropy loss. First, models are dedicated solely to the simplification task. 

\begin{table*}
    \centering
    \footnotesize
    \begin{tabular}{|p{1cm}|p{1cm}|p{10cm}|p{0.6cm}|}
    \hline
       \textbf{Token} & \textbf{Type} & \textbf{Text} & \textbf{Label}\\
    \hline
   \multirow{2}{1.5cm}{\textit{simplify:}} & \multirow{2}{1.5cm}{complex} & \multirow{2}{10cm}{People do it. So do chimpanzees, bonobos and baboons. Even dogs do it: They yawn when someone near them yawns.} & \multirow{2}{1.1cm}{-}  \\
       &  &  & \\
    \hline
   \multirow{2}{1.5cm}{\textit{read classify:}} & \multirow{2}{1.5cm}{gold-reference} & \multirow{2}{10cm}{When someone near you yawns, you yawn too. So do chimpanzees and baboons. Even dogs do it. They all seem to "catch" yawns.} & \multirow{2}{1.1cm}{4} \\
      &  &  & \\
    \hline
    \end{tabular}
    \caption{\new{Examples of inputs for the simplification models based on T5-family. We also include the readability label (`read classify:') for the gold reference.}}
    \label{eq:control_tokens_examples}
\end{table*}

\paragraph{Simplification Using Larger Datasets}
\label{simplification_larger}

Initially, we ran our experiments on professionally created corpora only, since automatically aligned datasets could introduce noise (e.g., poor simplifications) into the final results. Nevertheless, the existing datasets are relatively small, which also means that the models could fail to learn more aspects of language, such as fluency and grammar. 


We propose an alternative scenario in which we fine-tune the T5 models using Wikipedia-based corpora \note{and the default cross-entropy loss} as a first step. Therefore, the model will learn exclusively the simplification task with large texts (143,546 texts). Once the training is completed, we continue with the introduction of datasets with better quality to learn simplification, readability and coherence altogether, as proposed in previous sections. We evaluated this scenario only for the T5-large model, using the NewselaSL dataset in zero, few-shot and fine-tuning settings, as it showed the best performance in our preliminary results where training was limited to Newsela only (Section \ref{sec:results}).

\begin{table*}[tbpb] 
    \small
    \begin{center} 
    \begin{tabular}{|c|c|c|p{1.1cm}|p{1.1cm}|p{1.1cm}|p{1.1cm}|p{1.1cm}|p{1.2cm}|} \hline 
    \textbf{Dataset} & \textbf{Split} & \textbf{Pair} & \textbf{Lines} & \textbf{Sent/D} & \textbf{W/sent} & \textbf{Char/w} & $FRE\uparrow$ & $FKGL\downarrow$ \\
    \hline \multirow{6}{1.6cm}{Newsela-S} & \multirow{2}{*}{train} & source & 1528 & 48.25 & 23.51 & 5.19 & 46.37 & 10.90 \\
    \cline{3-9} & & target & 1528 & 98.11 & 17.50 & 5.09 & 87.52 & 3.30 \\
    \cline{2-3} \cline{3-9} & \multirow{2}{*}{valid} & source & 191 & 47.20 & 24.16 & 5.22 & 59.13 & 10.10 \\
    \cline{3-9} & & target & 191 & 97.76 & 17.69 & 5.11 & 94.76 & 2.60 \\
    \cline{2-3} \cline{3-9} & \multirow{2}{*}{test} & source & 191 & 47.59 & 23.65 & 5.17 & 53.75 & 12.20 \\
    \cline{3-9} & & target & 191 & 96.90 & 17.52 & 5.07 & 88.43 & 3.00 \\
    \cline{1-2} \cline{2-3} \cline{3-9} 
    \multirow{6}{1.6cm}{Newsela-SL} & \multirow{2}{*}{train} & source & 6089 & 48.11 & 23.66 & 5.20 & 60.95 & 9.40 \\
    \cline{3-9} & & target & 6089 & 97.75 & 19.72 & 5.14 & 74.08 & 6.40 \\
    \cline{2-3} \cline{3-9} & \multirow{2}{*}{valid} & source & 761 & 47.34 & 23.54 & 5.18 & 72.16 & 7.20 \\
    \cline{3-9} & & target & 761 & 96.66 & 19.63 & 5.13 & 76.22 & 5.60 \\
    \cline{2-3} \cline{3-9} & \multirow{2}{*}{test} & source & 762 & 48.65 & 23.10 & 5.17 & 61.50 & 11.30 \\
    \cline{3-9} & & target & 762 & 99.90 & 19.37 & 5.12 & 71.95 & 7.20 \\
    \cline{1-2} \cline{2-3} \cline{3-9} 
    \multirow{6}{1.6cm}{D-Wikipedia} & \multirow{2}{*}{train} & source & 132546 & 4.92 & 28.77 & 5.48 & 67.49 & 9.00 \\
    \cline{3-9} & & target & 132546 & 8.97 & 24.55 & 5.39 & 78.59 & 6.80 \\
    \cline{2-3} \cline{3-9} & \multirow{2}{*}{valid} & source & 3000 & 4.92 & 28.81 & 5.48 & 74.59 & 6.20 \\
    \cline{3-9} & & target & 3000 & 8.94 & 24.63 & 5.39 & 60.95 & 9.40 \\
    \cline{2-3} \cline{3-9} & \multirow{2}{*}{test} & source & 8000 & 5.01 & 28.84 & 5.49 & 67.69 & 8.90 \\
    \cline{3-9} & & target & 8000 & 9.10 & 24.66 & 5.40 & 83.66 & 4.80 \\
    \cline{1-2} \cline{2-3} \cline{3-9} \multirow{2}{1.6cm}{GCDC} & train & - & 4000 & 9.07 & 19.39 & 4.89 & 83.15 & 5.00 \\
    \cline{2-9} & test & - & 800 & 9.00 & 19.96 & 4.91 & 68.81 & 8.50 \\
    \hline
    \end{tabular} 
    \caption{Statistics for our selected datasets. We report the total number of documents (Lines), \note{average} number of sentences per document (Sent/D), \note{average} number of words per sentence (W/sent), \note{average} number of characters per word (Char/w) and readability indices FRE and FKGL.}
    \label{appendix:newsela_stats}
    \end{center} 
\end{table*}

\subsubsection{Readability-based TS Model}
\label{read_models}

The systems presented in Section \ref{simple_models} are limited to the simplification aspect. However, we suggest that with the addition of explicit readability levels, simplifications can be more varied and therefore, they could be better tailored to multiple audiences (e.g. non-native speakers, people with disabilities, and non-specialised audiences), similarly as it has been done in other fields such as summarisation \citep{luo-etal-2022-readability}. We explain the steps of the model architecture shown in Figure \ref{fig:combined_loss}, for simplification and readability. 
    \paragraph{\textbf{Simplification:}} complex texts are passed to the encoder of the T5 model, for the generation of a simple (simplified') candidate or prediction. We obtained the simplification loss ($loss_{simp}$) by comparing the simplification predictions and the gold-standard simplification.
    
    \paragraph{\textbf{Readability:}} after the simplification stage, simple texts (gold-standard simplifications) are passed to the T5-encoder, which generates a label prediction (readability label') according to the input simple sentence. For the readability loss ($loss_{read}$), we compared the predicted readability level of the gold-standard simplification with the gold-standard readability label. We expect that the readability labels range from 1 to 4, representing different degrees of readability. The classification task and the expected values are learnt during training.

In this model, we evaluate simplification and readability using the same encoder, together in the forward pass to obtain the corresponding loss. Finally, we sum these values together to calculate the total loss ($loss_{total}$) in Equation \ref{eq:multi_objective_loss}:

\begin{equation} \label{eq:multi_objective_loss}
 loss_{total} = loss_{simp} + loss_{read}
\end{equation}

Our combined loss considers the simplification aspects of the complex and simple pairs, but it also tailors simplification for a specific readability level, similar to a readability classification task \citep{vasquez-rodriguez-etal-2022-benchmark}. 
We control both tasks by the use of tokens to indicate the relevant task. For simplification generation we use the control token ``simplify:'' to obtain a simplified candidate. For the readability classification task, we control the task by using ``read classify:'' to obtain the level of simplification of the input sentence as explained in Section \ref{task_description}. In Table \ref{eq:control_tokens_examples} we show a template of how the input text is labelled for the simplification and readability tasks, respectively.


\subsubsection{Coherence-based Model}
\label{coherence_models}

As mentioned earlier, our main aim is the implementation of a simplification model considering readability, simplification and coherence. However, we are limited by the availability of annotated corpora that express the quality of a simplified text concerning these 3 elements simultaneously. For simplification and readability, we rely on the Newsela dataset as presented in Section \ref{read_models}. For coherence, we still do not have the annotation of coherence in this corpus. To alleviate this setting, we proposed the evaluation of coherence during training for each prediction. We intervene in the loss individually for each aspect (simplification, readability and coherence) and aggregated scores are reported after the final loss of each sample (i.e., each text) is calculated.\footnote{We refer to our preprocessing steps of our coherence models in the Appendix \ref{sec:data_preprocessing_ch9} and the selection steps of our coherence models in the Appendix \ref{appendix:model_selection}.} 



\begin{table*}[htbp]
    \centering
    \scriptsize
    \begin{tabular}{|p{3.6cm}|p{1cm}|p{1.8cm}|p{1.3cm}|p{1.3cm}|p{1.2cm}|p{0.9cm}|p{1cm}|}
        \hline
        \multirow{2}{1.1cm}{\textbf{Model}} & \multirow{2}{1.1cm}{\textbf{Setting}} & \multirow{2}{1.1cm}{\textbf{Train}} & \multirow{2}{1.1cm}{\textbf{Samples}} & \multirow{2}{1.1cm}{\textbf{Test}} & \multirow{2}{1.1cm}{\textbf{Samples}} &  \textbf{Fine-tuned} &  \multirow{2}{1.1cm}{\textbf{Task}} \\
        \hline
        NewselaS+simple & zero & $-$ & $-$ & NewselaS & 191 &  $\Box$ & $-$ \\
        \hline
        NewselaS+simple & few & NewselaS & 10 & NewselaS & 191 & \CheckedBox & S \\
        \hline
        NewselaS+simple & fine & NewselaS & 1528 & NewselaS & 191 & \CheckedBox & S \\
        \hline
        NewselaSL+simple & zero & $-$ & $-$ & NewselaSL & 762 & $\Box$ & $-$ \\
        \hline
        NewselaSL+simple & few &  NewselaSL & 10  &  NewselaSL  & 762 & \CheckedBox & S\\
        \hline
        NewselaSL+simple &fine &  NewselaSL & 6089 &  NewselaSL & 762 & \CheckedBox & S\\
        \hline
        \multirow{2}{3.5cm}{D-Wikipedia+NewselaSL +simple} &fine& D-Wikipedia, NewselaSL & 132546, 6089 & NewselaSL & 762  & \CheckedBox & S \\
        \hline
        NewselaSL+simple+read  & all &  NewselaSL & 6089 & NewselaSL & 762 & \CheckedBox & S+R \\
        \hline
        NewselaSL+simple+coh & all & NewselaSL & 6089 & NewselaSL & 762 & \CheckedBox & S+C \\
        \hline
        NewselaSL+simple+read+coh & all & NewselaSL & 6089 & NewselaSL & 762 & \CheckedBox & S+R+C \\
        \hline
        \multirow{2}{3.5cm}{D-Wiki+NewselaSL+ simple+read} & all & D-Wikipedia, NewselaSL & 132546, 6089 & NewselaSL & 762 &  \CheckedBox & S+R \\
        \hline
        \multirow{2}{3.5cm}{D-Wiki+NewselaSL+ simple+coh} &all& D-Wikipedia, NewselaSL & 132546, 6089 & NewselaSL & 762 &  \CheckedBox & S+C \\
        \hline
        \multirow{2}{3.5cm}{D-Wiki+NewselaSL+ simple+read+coh} & all& D-Wikipedia, NewselaSL & 132546, 6089 & NewselaSL & 762 &  \CheckedBox & S+R+C \\
        \hline
    \end{tabular}
    \caption{We present our proposed models according to the settings in Section \ref{task_description}. Each model is trained in zero-shot, few-shot, and fine-tuned, with a different number of samples. During training, we used either the standard CL loss for the TS task (S) standalone or a multi-task CL loss where we combined TS generation (S), readability classification (R) and/or coherence evaluation (C) at the same time. For the "all" setting, we report the parameters for the fine-tuned scenario, however, they were also evaluated in zero-shot and few-shot settings.}
    \label{tab:models_table}
\end{table*}


\subsection{Loss Function}
\label{sec:loss_function_by_sample}

In our experiments, we included simplification models using a default and a customised loss. In the default loss, we evaluate our predictions against the labels using the Cross-Entropy Loss (CL). This is the loss used in \note{our simplification models (Section \ref{simple_models})}, where we do not consider any readability and coherence aspect. Further, we used a customised loss for \textit{simplification+readability} model (Section \ref{read_models}), \textit{simplification+readability+coherence} and \textit{simplification+coherence} model variations.

In the customised loss simplification, readability and coherence aspects are considered individually for each sample. \new{Further, we perform the aggregation of the simplification loss, readability loss and coherence evaluation altogether. We calculate the total loss of the model as follows:}

\begin{enumerate}
    \item We calculate the CL loss between each prediction ($\hat{y}$) and its label ($y$) for the simplification model ($loss_{simp}(\hat{y}, y)$) and readability model ($loss_{read}(\hat{y}, y)$).
    \item We obtain the coherence value for the model prediction ($score_{coherence}(\hat{y})$).
    \item When a prediction is coherent ($score_{coherence}(\hat{y})$ is equal to 1), we reduce the loss ($loss_{simp}(\hat{y}, y)$+$loss_{read}(\hat{y})$) multiplying it by $\delta$, otherwise, we do not include any coherence evaluation in the loss as shown in Equation \ref{eq:multi_objective_loss_by_sample_1}:
    \begin{equation} \label{eq:multi_objective_loss_by_sample_1}
     \parbox{\dimexpr\linewidth}{
        $loss_{partial}(x) = \\
            \begin{cases}
                \delta\times (loss_{simp}(\hat{y}, y) + loss_{read}(\hat{y}, y)), & \\ \text{if } score_{coh}(\hat{y}) = 1 \\
                loss_{simp}(\hat{y}, y) + loss_{read}(\hat{y}, y), & \\ \text{otherwise}
            \end{cases}$}
    \end{equation}
    \item Finally, we present the average values for all the samples, as shown in Equation \ref{eq:multi_objective_loss_by_sample_2}:
    \begin{equation} \label{eq:multi_objective_loss_by_sample_2}
    \parbox{\dimexpr\linewidth-10em}{
    $loss_{total} = \frac{1}{n} \sum_{i=0}^{n} loss_{partial}(x_i)$
    }
    \end{equation}

\end{enumerate}

\subsection{Model Settings}
\label{appendix:model_settings}

We describe the proposed training settings for our models as follows:
 \paragraph{\textbf{Zero-shot}:} we directly evaluated the pre-trained model without using any training TS samples, a task which has not been seen before. This means that the model was not fine-tuned in the simplification generation and/or readability classification task. 
 \paragraph{\textbf{Few-Shot}:} we fine-tuned the model in simplification and/or readability task using just a few samples (10 samples) for 1 epoch. This will allow the models to learn the task and eventually, show a better evaluation in the test set of unseen data. We refer to this setting as a few-shot since we trained our models in a small set of instances.
 \paragraph{\textbf{Fine-tuning}:} the models are fine-tuned with the entire corpus for simplification generation and/or readability classification, further, we evaluate its performance on the test set.

\subsection{Data Selection and Preprocessing}
\label{ch8:data_selection}

To simplify text based on simplicity, readability, and coherence, it is necessary to have annotated datasets to perform effective model training. Firstly, we focused on the simplicity and readability aspects. Except for the OneStopCorpus, which is smaller in size, the Newsela dataset \citep{xu-etal-2015-problems} is the only corpus available with annotations for both simplicity and readability level.\footnote{We requested the English Newsela dataset for our experiments at \url{https://newsela.com/data/}.} We used the dataset in its original version \citep{xu-etal-2015-problems}, \textit{NewselaS}, and in an extended scenario, \textit{NewselaSL}. 

For the \textit{NewselaS}, we matched the complex documents, represented by level 0 and simple documents from level 4. In cases where level 4 documents were not available, we used the ones in level 3. 
For \textit{NewselaSL}, we considered adapting the datasets to include their simplification degree as well. We created a pair for each complex article and its corresponding simple article, one per each simplification level. Hence, the complex article (level 0), will be mapped with its simpler counterparts (1, 2, 3 and 4 when available). Each record will include the following features: source and target simplification and simplification level of the target simplification (henceforth, readability label). 

Additionally, \noteb{we use GCDC---a professionally annotated} dataset \citep{lai-tetreault-2018-discourse} for training the coherence models as an additional element in the simplification model and D-Wikipedia \citep{sun-etal-2021-document} 
as a means to initially fine-tune language models for better fluency.\footnote{We include the datasets statistics in Table \ref{appendix:newsela_stats} and additional details in the Appendix  \ref{sec:datasets_coherence}.}

Our rationale for the proposed pipeline relies on the limitations of the Newsela dataset as sentences in documents are not aligned between complex and simple articles. Also, some of them are too large to fit together into the model. Therefore, we decided to build documents of 10 sentences each, with the hypothesis that the first part of the article would be aligned. We followed this decision as well for our manual analysis, as we discuss in Section \ref{ch9:discussion}.

\section{Experiments}
\label{experiments}

We discuss the implementation of the T5-models in zero-shot, few-shot, and fine-tuning scenarios, including further customisation of models for readability and coherence. Further, we explain the datasets selection (Section \ref{ch6:datasets}), model implementation (Section \ref{models_implementation}) and evaluation (Section \ref{ch8:evaluation}).

\subsection{Datasets}
\label{ch6:datasets}

For our experiments, we have selected English Newsela corpus as the most suitable resource, given their annotations on simplification and readability. This corpus has 1528 articles, written in 4 levels of readability, where 0 represents the complex version and 4 the simplest one. As described in Section \ref{ch8:data_selection}, we created our splits for train, validation and test. Although the corpus is not large, it is enough for the implementation of a multi-task loss proposed in Section \ref{read_models}. Despite the availability of other document-level simplification datasets such as D-Wikipedia, they do not have any readability or coherence annotations. For coherence, we used the GCDC corpus, a collection of 1200 texts for each domain (1000 for training and 200 for testing). This corpus is in the domain of emails and business reviews, professionally annotated by experts. We use this dataset to fine-tune our models for coherence evaluation. \noteb{For the coherence annotations in this dataset, we use the consensus label defined by the experts, as their judgement was more accurate than the crowd-sourced labels.} Finally, we evaluated all of our experiments using \note{a model fine-tuned on the }D-Wikipedia datasets, for the simplification task.

\subsection{Models}
\label{models_implementation}

For the implementation of our models in Section \ref{sec:models_selection}, we used the pretrained language models T5-small\footnote{\url{https://huggingface.co/t5-small}} and T5-large\footnote{\url{https://huggingface.co/t5-large}} from the Hugging Face using the Pytorch Lightning framework.\footnote{\url{https://lightning.ai/pages/open-source/}} We summarise in Table \ref{tab:models_table} the proposed models for our task.\footnote{We include our training details in the Appendix \ref{appendix:training_details}.}

\begin{table*}[htpb!]\centering
\scriptsize
\begin{tabular}{|p{1.6cm}|p{1.5cm}|p{1cm}|p{0.9cm}|p{1.2cm}|p{1.2cm}|p{1.2cm}|p{1cm}|p{1cm}|p{1cm}|}
\hline
\textbf{Model}  &\textbf{Dataset} &\textbf{Loss} &\textbf{Setting} &\textbf{$\operatorname{D-SARI_{S}}\uparrow$} &\textbf{$FKGL_{C}\downarrow$} &\textbf{$FKGL_{S}\downarrow$} &\textbf{$FRE_{C}\uparrow$} &\textbf{$FRE_{S}\uparrow$} &\textbf{$COH_{S}\uparrow$} \\
\cline{1-10}
muss\_mined & \multirow{2}{2cm}{-} & \multirow{2}{2cm}{-} & \multirow{2}{2cm}{-}  & 19.231 & \multirow{2}{*}{9.758} & 8.301 & \multirow{2}{*}{60.97} & 66.709 & 0.739 \\
\cline{1-1} 
muss\_wl &  &  & & 19.993 &  & 6.921 &  & 72.248 & 0.738 \\
\hline
\multirow{3}{*}{t5-small} & \multirow{6}{*}{NewselaS} &\multirow{3}{*}{simple} & zero & 32.271 &\multirow{6}{*}{10.065} &5.019 &\multirow{6}{*}{60.049} &81.170 &0.131 \\
\cline{4-4}
& & & few & 32.778 & &4.990 & &81.391 &0.120 \\
\cline{4-4}
& & & fine &\textbf{44.935} & &\textbf{2.942} & &\textbf{89.231} &0.000 \\
\cline{1-1} \cline{3-4}\cline{7-7}\cline{9-10}
\multirow{3}{*}{t5-large}  & & \multirow{3}{*}{simple} & zero & 30.531 & &5.920 & &77.924 &0.204 \\
\cline{4-4}
& & & few & 30.639 & &5.713 & &78.593 &0.209 \\
\cline{4-4}
& & & fine & \textbf{50.089} & &\textbf{3.238} & &\textbf{87.753} &0.000 \\
\hline
\end{tabular}
\caption{Our results on \textit{NewselaS} \note{systems} including D-SARI scores. We also report the FKGL scores for the complex ($FKGL_{C}$) and simple ($FKGL_{S}$) text. Similarly, we report the FRE scores for the complex ($FRE_{C}$) and simple ($FRE_{S}$) text as well. For each setting (e.g., simple) we report the results for zero, few, and fine-tuned.}
\label{results_newselaS}
\end{table*}

\subsection{Evaluation}
\label{ch8:evaluation}

For the automatic evaluation of the results, we use D-SARI (SARI metric extended for document-level TS) for simplification and FKGL and FRE for readability. \notec{We understand the limitations of existing readability evaluation metrics \citep{tanprasert-kauchak-2021-flesch}. However, we used these metrics as a means to compare with previous work and as a complementary evaluation metric to measure simplification generation quality}.

Additionally, we performed a manual analysis of predictions to have a further understanding of the model results. We selected the following samples for the analysis, which we further discuss on Section \ref{ch9:discussion}:
\begin{itemize}
    \item The \textit{NewselaS} test set, evaluated on the fine-tuned \textit{T5-large} model comparing the complex, prediction and the gold standard for a total of 191 triplets (573 texts);
    \item The \textit{NewselaSL} test set, evaluated on the fine-tuned \textit{T5-large} model, comparing the complex, gold standard and predictions of the \textit{NewselaSL+simple+fine}, \textit{NewselaSL+simple+read}, \textit{NewselaSL+simple+coh} and \textit{NewselaSL+simple+read +coh} model;
    \item The \textit{NewselaSL} test set, evaluated on the fine-tuned \textit{T5-large} model, comparing the complex, gold standard and predictions of the \textit{D-Wiki+NewselaSL+simple+fine}, \textit{D-Wiki+NewselaSL+simple+read}, \textit{D-Wiki+NewselaSL+simple+coh} and \textit{D-Wiki +NewselaSL+simple+read+coh} model.
\end{itemize}


\section{Results}
\label{sec:results}

We present our results for the model using \textit{NewselaS} in Table \ref{results_newselaS}. For both \textit{T5-small} and \textit{T5-large}, the zero-shot task showed the lowest performance. However, these models consistently gave a better performance from zero-shot setting to fine-tuning for D-SARI scores. For the D-SARI and FKGL metric, the difference between the \textit{zero-shot} and the \textit{few-shot} scenario was small, and in general, both models had the best performance in \textit{fine-tuning} for all metrics. We can observe similar behaviour for the FRE metric. Overall, the readability of the outputs was improved concerning the complex \new{documents} for both FKGL and FRE metrics. \note{Also, all the models reported better scores than the presented baselines, \textit{muss\_en\_mined} and \textit{muss\_en\_wikilarge\_mined}.}\footnote{For more details, see Appendix \ref{appendix:baseline_models}.}

For the \textit{NewselaSL} dataset, we obtained lower results in the \textit{zero-shot} setting, compared to the performance of \textit{NewselaS}, as shown in Table \ref{results_newselaSL_t5_large}. \note{However, for the \textit{fine-tuning} settings, we obtained better D-SARI scores in both \textit{T5-small} and \textit{T5-large}. This is not the case for the \textit{few-shot} scenario, as the improvement to the \textit{zero-shot} scenario was minor.} In the scenario where simplification models are fine-tuned with automatically aligned corpora using D-Wikipedia, the results in \textit{zero-shot} and \textit{few-shot} are significantly larger than the previous setting (no fine-tuning in any simplification corpora) with an increase of 29.25 and 30.01 (using the average for all loss types) in SARI score, for \textit{zero-shot} and \textit{few-shot} respectively. \note{In this dataset, the results for the \textit{zero-shot} and \textit{few-shot} scenarios using \textit{t5-large} were closely equivalent to the \textit{muss\_en\_mined} baseline. For the \textit{t5-small} model, all scenarios improved the results presented by both MUSS baselines.}

In contrast, the model trained in \textit{NewselaSL} dataset with \textit{simple+read+coh} and \textit{simple+coh}, showed a decrease in performance in comparison to the rest of the experiments (\textit{simple} and \textit{simple+read}), but still better than the experiments in \textit{zero-shot} and \textit{few-shot} setting. For the \textit{t5-large} model trained in \textit{D-Wiki+NewselaSL} using \textit{simple+read+coh}, there was a small increase of 0.43 with respect to the D-SARI metric, in comparison to the \textit{simple+read} scenario for the same dataset. Overall, the predictions showed better readability in all settings than the complex texts. Similar to the experiments in \textit{t5-small}, \textit{simple+read+coh} also showed a decrease in both SARI and similar results in the readability metrics concerning the \textit{simple} setting. For \textit{D-Wiki+NewselaSL}, we noticed a minor decrease in D-SARI score in \textit{simple+coh} compared to \textit{simple+read+coh} scenario. The fine-tuned model using \textit{D-Wikipedia} and \textit{NewselaSL} reported scores are better in all scenarios.

\section{Discussion}
\label{ch9:discussion}

Firstly, we discuss the results of the \textit{zero-shot} scenario. Although the \textit{t5-model} was previously trained in multiple text-to-text tasks such as question answering, summarisation, and classification, it was not explicitly trained in simplification generation (``simplify:'') and readability classification (``read classify:'') task. Therefore, the D-SARI performance in \textit{NewselaS} zero-shot experiments was low with a SARI score of 32.271 and 30.531, especially in \textit{NewselaSL} using a \textit{simple} loss, with values of 24.813 and 22.690 for \textit{t5-small} and \textit{t5-large}, respectively. 
In \textit{NewselaS}, complex text (level 0) and simple texts (level 4) are significantly different from each other since the pairs are not formed of contiguous readability levels. In \textit{NewselaSL}, more complexity is added with the granularity on the readability levels, which we used to distinguish between the different levels of simplifications. For the D-Wikipedia experiments, the models explicitly learn the simplification task, showing a major improvement when evaluated in a \textit{zero-shot} fashion using the \textit{NewselaSL} dataset. 

Secondly, we present the analysis for the \textit{few-shot} scenario in Table~\ref{results_newselaS} and Table \ref{results_newselaSL_t5_large}. \note{We consider} that since the task of text simplification and readability classification are similar to previous pretrained tasks (i.e., summarisation and sentiment classification tasks, respectively) in the model, it should be enough to train the model in a few samples to learn the new task. In this setting, we showed an increase of 0.26 on average for the \textit{few-shot}, in comparison to the \textit{zero-shot} experiments. Hence, by training with few samples (10 samples in this case) and running a minimum training epoch (1 epoch in this case), control token-based models could learn better the proposed task. However, it is necessary to train with larger samples for a more significant performance increase.

For the multi-task loss setting (\textit{simple+read} and \textit{simple+read+coh}), both simplification and readability tasks are learned in parallel from the first training steps. The performance measured by D-SARI in a few-shot setting is also lower, and more data and training iterations are required to stabilise the loss (i.e., learn properly each task), as seen in \textit{fine} experiments. In the models fine-tuned with D-Wikipedia, the \textit{few-shot} setting shows a larger increase concerning the \textit{zero-shot} setting, as it has already seen the simplification task using the same control tokens (i.e., ``simplify:'').

Thirdly, we look further into the \textit{fine-tuning} experiments. For both datasets, \textit{NewselaS} and \textit{NewselaSL} we obtain the best D-SARI score within all of our settings \textit{zero-shot}, \textit{few-shot} and \textit{fine-tuning}. Within the multi-task loss, the addition of readability to the loss (\textit{simple+read}) showed a decrease of 0.78 for \textit{t5-small} and 1.274 for the \textit{t5-large} model, both using the \textit{NewselaSL} dataset. For \textit{D-Wiki+NewselaSL}, readability experiments showed a minor decrease of 0.876, which could be related to the fact that the readability classification task has not been seen before, in contrast to the scenario in which simplification and readability tasks were both new to the model. 

\begin{table*}[!htp]\centering
\footnotesize
\begin{tabular}{|p{1.6cm}|p{1.5cm}|p{1cm}|p{1cm}|p{1.2cm}|p{1.2cm}|p{1.2cm}|p{1cm}|p{1cm}|p{1cm}|}
\hline

\textbf{Model} &\textbf{Train} &\textbf{Loss} &\textbf{Setting} &\textbf{$\operatorname{D-SARI_{S}}\uparrow$} &\textbf{$FKGL_{C}\downarrow$} &\textbf{$FKGL_{S}\downarrow$} &\textbf{$FRE_{C}\uparrow$} &\textbf{$FRE_{S}\uparrow$} &\textbf{$COH_{S}\uparrow$} \\
\cline{1-10}

muss\_mined & \multirow{2}{1.65cm}{-} & \multirow{2}{1cm}{-} & \multirow{2}{1cm}{-} & 22.506 & \multirow{2}{*}{9.451} & 8.068 & \multirow{2}{*}{62.045} & 67.556 & 0.729 \\
\cline{1-1} 
muss\_wikiL &  &  & & 23.766 &  & 6.727 &  & 72.830 & 0.732 \\
\hline
    \multirow{12}{*}{t5-small} &\multirow{12}{*}{NewselaSL} &\multirow{3}{1cm}{simple} &zero &24.813 &\multirow{12}{*}{9.899} &6.756 &\multirow{12}{*}{60.120} &73.680 &\textbf{0.307} \\
    & & &few &25.253 & &6.599 & &74.591 &0.249  \\
    & & & fine &\textbf{48.971} & &\textbf{5.787} & &\textbf{76.592} &0.026 \\
    \cline{3-4} \cline{7-7} \cline{9-10}
    & &\multirow{3}{1cm}{simple +read} &zero&24.813 & &6.756 & &73.680 &\textbf{0.307} \\
    & & & few &25.208 & &6.605 & &74.531 &0.260 \\
    & & & fine &\textbf{47.697} & &\textbf{5.798} & &\textbf{76.496} &0.020 \\
    \cline{3-4} \cline{7-7} \cline{9-10}
    & &\multirow{3}{1cm}{simple +coh} & zero &24.813 & &6.756 & &73.680 &\textbf{0.307} \\
    & & & few &25.257 & &6.604 & &74.566 &0.251 \\
    & & & fine &\textbf{47.408} & &\textbf{5.722} & &\textbf{76.825} &0.026 \\
    \cline{3-4} \cline{7-7} \cline{9-10}
    & &\multirow{3}{1cm}{simple +read +coh} & zero &24.813 & &6.756 & &73.680 &\textbf{0.307} \\
    & & & few &25.209 & &6.608 & &74.517 &0.259 \\
    & & & fine &\textbf{46.533} & &\textbf{5.812} & &\textbf{76.535} &0.022 \\
    \hline
\multirow{24}{*}{t5-large} &\multirow{12}{*}{NewselaSL} &\multirow{3}{*}{simple} & zero & 22.690 & &7.697 &  &70.026 &\textbf{0.429} \\
\cline{4-4}
& & & few & 22.821 & &7.425 & &70.799 &0.416 \\
\cline{4-4}
& & & fine & \textbf{53.400} & &\textbf{5.720} & &\textbf{76.411} &0.041 \\
\cline{4-4}
\cline{3-4} \cline{7-7} \cline{9-10}
& &\multirow{3}{1cm}{simple +read} & zero & 22.690 & &7.697 & &70.026 &\textbf{0.429} \\
\cline{4-4}
& & & few & 22.745 & &7.744 & &69.926 &0.403 \\
\cline{4-4}
& & & fine & \textbf{52.619} & 9.899 &\textbf{5.705} & 60.120 &\textbf{76.536} &0.038 \\
\cline{3-4} \cline{7-7} \cline{9-10}
& &\multirow{3}{1cm}{simple +coh} & zero & 22.690 & &7.697 & &70.026 &\textbf{0.429} \\
\cline{4-4}
& & & few & 22.820 & &7.375 & &70.993 &0.417 \\
\cline{4-4}
& & & fine & \textbf{51.519} & &\textbf{5.729} & &\textbf{76.486} &0.041 \\
\cline{3-4} \cline{7-7} \cline{9-10}
& &\multirow{3}{1cm}{simple +read +coh} & zero & 22.690 & &7.697 & &70.026 &\textbf{0.429} \\
\cline{4-4}
& & & few & 22.751 & &7.730 & &69.985 &0.411 \\
\cline{4-4}
& & & fine & \textbf{51.057} & &\textbf{5.747} & &\textbf{76.452} &0.034 \\
\cline{2-10}
&\multirow{12}{1.5cm}{D-Wiki+ NewselaSL} &\multirow{3}{*}{simple} & zero & 51.935 & &6.752 &  &74.159 &0.024 \\
\cline{4-4}
& & & few & 53.896 & &6.468 & &74.782 &0.031 \\
\cline{4-4}
& & & fine& \textbf{55.441} & &\textbf{5.693} & &\textbf{76.452} &\textbf{0.049} \\
\cline{3-4} \cline{7-7} \cline{9-10}
& &\multirow{3}{1cm}{simple +read} & zero & 51.935 & &6.752 & &74.159 &0.024 \\
\cline{4-4}
& & & few & 52.221 & &6.779 & &74.073 &0.028 \\
\cline{4-4}
& & & fine & \textbf{54.565} & 9.899 &\textbf{5.685} & 60.120 &\textbf{76.505} &\textbf{0.046} \\
\cline{3-4} \cline{7-7} \cline{9-10}
& &\multirow{3}{1cm}{simple +coh} & zero & 51.935 &  &6.752 & &74.159 &0.024 \\
\cline{4-4}
& & & few & 53.919 & &6.466 & &74.797 &0.031 \\
\cline{4-4}
& & & fine & \textbf{54.398} & &\textbf{5.690} & &\textbf{76.538} &\textbf{0.046} \\
\cline{3-4} \cline{7-7} \cline{9-10}
& &\multirow{3}{1cm}{simple +read +coh} & zero & 51.935 &  &6.752 & &74.159 &0.024 \\
\cline{4-4}
& & & few & 52.219 & &6.779 & &74.072 &0.028 \\
\cline{4-4}
& & & fine & \textbf{54.993} & &\textbf{5.688} & &\textbf{76.496} &\textbf{0.047} \\
\hline
\end{tabular}
\caption{\new{Our results on the \textit{NewselaSL} dataset for t5-small, t5-large model and baselines. We report $\operatorname{D-SARI_{S}}$, coherence ($COH_{S}$) for simple texts and FKGL and FRE scores for the complex ($FKGL_{C}$, $FRE_{C}$) and simple ($FKGL_{S}$, $FRE_{S}$) texts. For each setting (e.g., simple, simple+read) we report the results for zero, few and fine-tuned model.}}
\label{results_newselaSL_t5_large}
\end{table*}

For the readability metrics, both FKGL and FRE showed similar results. Although the overall increase in this setting was not as large as in other experiments, it shows an indication that both tasks can be stabilised together and that readability could also support better simplifications. Simplifications should be considered for multiple audiences, meaning the same complex text should have several levels of readability. Our proposed \textit{NewselaSL} dataset searches to model this scenario. 

Additionally, we analyse the use of coherence in our experiments during training. For the models trained using the \textit{NewselaSL} dataset, the combined loss \textit{simple+read+coh} showed a decrease of 1.164 for the \textit{t5-small} model and of 1.56 in \textit{large} in SARI score, in comparison to \textit{simple+read} model. We observe a similar scenario for \textit{simple+coh}, where the fine-tuned scores are lower than \textit{simple} and \textit{simple+read} for both models. However, where the model uses a larger dataset for fine-tuning (\textit{D-Wiki+NewselaSL}), \textit{simple+read+coh} showed a slight improvement of 0.595 compared to the previous setting for \textit{simple+coh}. We hypothesise that at the start of the training stage, predictions could be incoherent, so the loss would not benefit from coherence in the initial steps. 



Finally, we consider the results from the manual analysis. We performed a comparison between 36 texts extracted from our articles (complex, gold references and predictions), using approximately 10 sentences of each text as explained in Section \ref{ch8:data_selection}.\new{\footnote{Analysis was performed by the first author of this paper.}} 
In terms of simplification operations, the model mainly performs \textbf{lexical simplification} and \textit{syntactic simplification}, focused on sentences split, deletion to create shorter sentences and paraphrasing (See example \ref{tab:lexical_syntactic_example_1}, \ref{tab:lexical_syntactic_example_2} and \ref{tab:lexical_syntactic_example_3} in the Appendix).\footnote{Due to data licensing restrictions, we do not show large samples of generated texts from the Newsela corpus.}

These features can also be seen in the gold-standard simplifications, where the content is significantly reduced from the original articles. Overall, the simplifications from all our models are fairly similar, with minor variations on the lexical and syntactic aspects. We acknowledge that the main limitation of our results is the fluency (i.e., grammaticality), meaning preservation and coherence of the text. Although the models were trained with quality data (i.e., professionally annotated), it is not enough to generate sentences that are both grammatical but also simplified. Also, we found issues in the coherence of the text.
We believe that additional stages of fine-tuning with more simplified data and larger models could improve our results. The use of readability and coherence in simplification is encouraging, however, it is necessary to evaluate further scenarios such as assigning different weights for each aspect in simplification and initial warm-up steps for each task to proficiently manage all proposed tasks.  In regards to our manual analysis, we acknowledge that while our findings are insightful for the community, it is still not clear the distinction between the benefits of coherence and readability independently. A more detailed and extensive human evaluation would need more specialised people and more diverse outputs to achieve conclusive results from this process. We discuss further the limitations and future work in Section \ref{ch9:future_work}. 

Finally, we conclude that the inclusion of coherence and readability into simplification is not solved. This direction is underexplored, given that coherence is difficult and subjective to the model. We chose this avenue to share the possible alternatives and we find it valuable to share our findings until the moment. We expect that our research motivate the community to continue to tackle this direction effectively, with the development of better datasets and methods, as we will discuss Section \ref{ch9:future_work}.

\section{Limitations}
\label{ch9:future_work}

In this section, we discuss the limitations and future work for the research in document-level text simplification:

    \paragraph{\textbf{Data resources}:} for the development of better coherence models, it is necessary to create larger and more varied datasets, professionally annotated within multiple domains. This will contribute to more precise coherence models for the domain understudy. For simplification, specifically for the Newsela dataset, complex articles are usually large in comparison to their simple counterpart. Since these are not strictly aligned\note{, (i.e., sentence-by-sentence simplification)}, it is difficult to learn explicit simplification operations beyond those already observed in this study.
    \paragraph{\textbf{Methods for simplification}:} simplifications will vary according to the target audience, therefore, it would also be interesting to simplify documents according to their domain, using customised methods that classify data with more granularity (e.g., topic-based). Also, using datasets at a document level available in other languages could be beneficial. Similarly to experiments in previous work \cite{vasquez-rodriguez-etal-2023-document},
    the model could be improved with an initial fine-tuning stage using a cross-lingual setting, leveraging simplification resources at a document level that are available in other languages.
    \paragraph{\textbf{Models for coherence}:} for our coherence model training stage, we used the professionally annotated GCDC corpus. However, we still depend on the generalisation of the model in other domains, such as news for Newsela dataset. This model was proposed as a guide for simplification generation, however, it should be improved further, together with the datasets enhancements proposed in \textit{Data resources}.
    \paragraph{\textbf{Hyper-parameters tuning}:} we chose our prompts ``simplify'' and ``read classify'' as a starting point for using t5-models for simplification. However, further engineering of prompts could enable better performance of the models as done in previous work \citep{aumiller-gertz-2022-unihd,vasquez-rodriguez-etal-2022-uom}.
    \new{Finally, depending on the selected task, simplification, readability and coherence could be weighted to achieve a personalisation step towards the target audience, as all the aspects may not be equally relevant during training.}
    \paragraph{\textbf{Computing resources}:} an alternative method to improve the quality and the fluency of the simplifications, it could be possible to train on larger language models (i.e., \textit{t5-3b}, \textit{t5-11b}) and larger datasets.

\bibliography{main}

\appendix

\section{Datasets}
\label{sec:datasets_coherence}
     \paragraph{\textbf{Grammarly Corpus of Discourse Coherence (GCDC)}} \citep{lai-tetreault-2018-discourse}: a collection of 1200 texts for each domain where 1000 samples are used for training and 200 for testing. These domains are based on the Yahoo Questions and Answers dataset,\footnote{\url{https://webscope.sandbox.yahoo.com/catalog.php?datatype=l}} emails from Hillary Clinton’s office,\footnote{\url{https://foia.state.gov/Search/Results.aspx?collection=Clinton_Email}} emails from the Enron Corpus\footnote{\url{https://www.cs.cmu.edu/~./enron/}} and business reviews from the Yelp Open Dataset.\footnote{\url{https://www.yelp.com/dataset}} \noteb{This dataset was annotated by 13 professional annotators and 32 lay people to evaluate the performance of untrained people. Among these groups, the experts had previous annotation experience, while the non-professional annotators were crowdsourced through Mechanical Turk,\footnote{\url{https://www.mturk.com/}} filtered by a qualification exam. The annotators were given an initial definition of coherence and examples for each category to get familiarised with the task. Further, each text was rated on a 3-point scale with values from 1 to 3 to denote low, medium and high coherence. Each text is rated by 3 experts and 5 crowdsourcing annotators. These annotations were released, including the individual score per each rater and a consensus label for each group. The consensus label was calculated based on the mean coherence value, using a threshold for a 3-way classification as follows: low $\leq$ 1.8 $<$ medium $\leq$ 2.2 $<$ high. For the annotator agreement between raters, the mean values for the intraclass correlation (ICC) and quadratic weighted Cohen’s $k$ were reported.}\footnote{The GCDC dataset is available upon request to the authors.}
     \paragraph{\textbf{D-Wikipedia}} \citep{sun-etal-2021-document}: this corpus was proposed as the first Wikipedia-based corpora for simplification at a document-level with 143,546 texts using Wikipedia\footnote{\url{https://www.wikipedia.org/}} and Simple Wikipedia\footnote{\url{https://simple.wikipedia.org/wiki/Main_Page}} article headers of each article. This refers to the main text after the main title, which is also referred to as ``abstracts'' by the authors. The alignment of the complex and simple Wikipedia was done automatically, therefore, the dataset is prone to some errors (e.g., incorrect alignments). However, the dataset can be relevant since it is large and suitable for learning the task of simplification, despite its known limitations \citep{cripwell-etal-2023-document}.

\section{Models}

\subsection{Training Details}
\label{appendix:training_details}

We trained the models using one Nvidia A100 GPUs with 80GB GPU RAM. For our models' hyperparameters, we used a learning rate of $2\times 10^{-5}$, weight decay of 0.01, batch size of 8, an AdamW optimizer \citep{Loshchilov_2019} and 5 training epochs, except for the few-shot models which were trained only for 1 epoch. For $\delta$, we selected a value of 0.90 based on the impact it would have on the loss values for readability and simplification. Further tuning of these parameters will remain as future work. 

\subsection{Coherence Model Selection}
\label{appendix:model_selection}

\note{For the evaluation of coherence in simplification, we assessed 3 different systems: a BERT-based baseline \citep{devlin-etal-2019-bert}, sentence-embedding \citep{reimers-gurevych-2020-making} with an LSTM \citep{Graves-2005} and the SetFit model \citep{Tunstall-2022}. For this benchmark, we included scenarios in which data is synthetically generated, as has been traditionally done in previous work \citep{li-hovy-2014-model, li-jurafsky-2017-neural, xu-etal-2019-cross}. We have also included the evaluation of real-world scenarios \citep{lai-tetreault-2018-discourse}, which will determine the final selection of the model for TS evaluation.\footnote{Although we also experimented with synthetic data, we limit the discussion of our results to human-annotated data as this would be our main target datasets for simplification.}}

\subsubsection{BERT-based model}

\note{We considered a \textbf{BERT}-based \citep{devlin-etal-2019-bert} model as a baseline, particularly BERT base (uncased), with no further modifications to its parameters. In this model, we aim to contrast the input data representations encoded by BERT, instead of sentence embeddings (such as by sBERT \citep{reimers-gurevych-2020-making}).}
\note{We hypothesise that BERT embeddings would fail to capture the logical relationships between sentences if any, and that the model will under-perform in comparison to the ones which use better data representations such as sentence embeddings. Our main target is the evaluation at a discourse level and not at a word level, however, we include the BERT model for comparison. Finally, we fine-tune the BERT model using the preprocessed texts for the classification of coherence.}

\begin{figure}[thpb]
\centering

\includegraphics[width=0.95\linewidth]{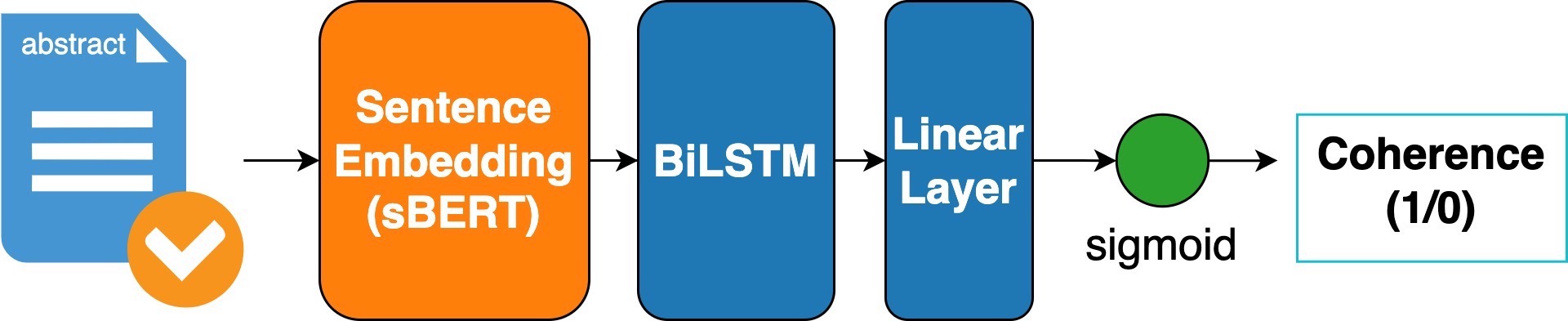}
\caption{SBERT-based model architecture}
\label{fig:sbert_model}
\end{figure}

\subsubsection{sBERT+LSTM-based model}
\note{Next, we considered our system \textbf{sBERT+LSTM}. In terms of data representation, most of the previous work (refer to related work in Appendix \ref{sec:related_work}) encodes text as strings or token-based embedding. Hence, text is encoded individually per word without capturing the semantic meaning of these words at a sentence or document level altogether. To mitigate this issue, we proposed a model that uses sentence transformers (sBERT) as inputs.} \note{In Figure \ref{fig:sbert_model}, we present the sBERT+LSTM model, which describes the architecture of our system. We preprocessed our datasets as described in Section \ref{sec:data_preprocessing_ch9}. Then, the inputs are encoded using an sBERT model. Each encoded text (i.e., instance) is represented by a collection of 10 encoded sentences. Texts are passed to a biLSTM layer and a linear layer \citep{Graves-2005}. Finally, we performed binary classification of our inputs using the sigmoid function which will determine if the text is coherent (1) or incoherent (0).}


\begin{figure}[thpb]
\centering
\includegraphics[width=0.90\linewidth]{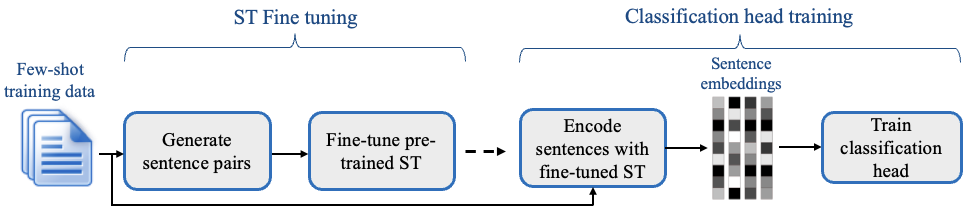}
\caption{SetFit model architecture \citep{Tunstall-2022}. ST stands for Sentence Transformers.}
\label{fig:setFit_model}
\end{figure}

\subsubsection{SetFit model}

Finally, we considered the \textbf{SetFit model}. Recently, \citet{Tunstall-2022} proposed this model, which facilitates the implementation of the data augmentation step, typically challenging for contrastive learning methods \citep{Jaiswal-2021}. In this architecture (Figure \ref{fig:setFit_model}), sentence embeddings are used together with contrastive learning in a few-shot manner, implementing internally the required data representation. As a first step, sentence embeddings are fine-tuned using a few samples from the dataset. Then, the fine-tuned sentence transformer is used to encode the initial input sentences for training the model. This model does not expect large datasets to get competitive results, since it was designed to work in a few-shot setting. We preprocessed our input data concatenated in a single string, as in the BERT model as we expect one coherence value per each text. Once we prepared our data, we fine-tuned the SetFit model for coherence classification.

\note{For the experiments using the GCDC dataset, we considered the three aforementioned models. In this setting, the SetFit model showed the best performance. 
Therefore, we selected this model for our simplification system. In contrast, \textit{sBERT-LSTM} and \textit{BERT}, did not achieve comparable performance, as in the synthetic data scenarios.}

\section{Coherence Data Selection}

\subsection{Data Preprocessing}
\label{sec:data_preprocessing_ch9}

We selected texts (Section \ref{ch8:data_selection}) from the general domain (e.g., news, emails and business reviews) for our coherence evaluation task. Then, we preprocessed the text to standardise it among all models as follows: 1) We performed sentence segmentation on the input text using the NLTK library \citep{bird2009natural}; 2) We selected the first 10 sentences of each instance since texts can have a varied number of sentences. These sentences were extracted from the first paragraph, which is likely to make sense by itself, without needing the following context. We also considered splitting the articles into multiple paragraphs, however, there is no guarantee that the consequent paragraphs are connected when they are extracted from their context, causing disruptions in the narrative; 3) We add one padding token per sentence to our tokenised text to those that have less than 10 sentences to complete the remaining sentences.

\newpage
\begin{table*}
    \centering
    \begin{tabular}{|p{4.4cm}|p{9cm}|}
        \hline
        \textbf{Type} & \textbf{Text} \\
        \hline
        \textit{Complex:}  &  \multirow{2}{9cm}{...The sighting of the red fox - one of 14 \textbf{mammals} \textit{protected by} California -...} \\
        & \\
        \hline
        \textit{Gold-reference:} & \multirow{2}{9cm}{ ...California protects 14 \textbf{animals}. The red fox is one of them. \textit{It is against the law to hunt or kill them}...} \\
        & \\
        \hline
         \multirow{2}{4.4cm}{\textit{Simple\phantom{e}(NewselaS+simp +fine):}}  &  \multirow{2}{9cm}{...The's 14 \textbf{animals}. The red fox was one of them. \textit{The is one the law to hunt wild animals...}} \\
         & \\
         \hline
    \end{tabular}
    \caption{Example 1: Manual analysis for simplification models using NewselaS.}
    \label{tab:lexical_syntactic_example_1}
\end{table*}

\begin{table*}
    \centering
    \begin{tabular}{|p{4.4cm}|p{9cm}|}
        \hline
        \textbf{Type} & \textbf{Text} \\
        \hline
        \textit{Complex:}  &  \multirow{2}{9cm}{...Volkswagen has admitted installing ``defeat devices" in as many as 11 million diesel engines... } \\
        & \\
        \hline
        \textit{Gold-reference:} & \multirow{2}{9cm}{...The automaker \textbf{cheated} emissions tests on their diesel cars...} \\
        & \\
        \hline
         \multirow{2}{4.4cm}{\textit{Simple\phantom{e}(NewselaSL+simp +read)}}  &  \multirow{2}{9cm}{...The dealers up devices to \textbf{make} the cars information tests...} \\
         & \\
         \hline
         \multirow{2}{4.4cm}{\textit{Simple\phantom{e}(NewselaSL+simp +read+coh):}}  &  \multirow{2}{9cm}{...The dealers up cars to \textbf{make} the cars information tests...} \\
         & \\
         \hline
    \end{tabular}
    \caption{Example 2: Manual analysis for simplification models using NewselaSL.}
    \label{tab:lexical_syntactic_example_2}
\end{table*}

\begin{table*}
    \centering
    \begin{tabular}{|p{4.7cm}|p{8.5cm}|}
        \hline
        \textbf{Type} & \textbf{Text} \\
        \hline
        \textit{Complex:}  &  \multirow{2}{8.5cm}{...Named for the color of their fatty tissue, green turtles go about nesting in a \textbf{peculiar} way...} \\
        & \\
        \hline
        \textit{Gold-reference:} & \multirow{2}{8.5cm}{...They knew that green turtles go about nesting in a \textbf{peculiar way}... } \\
        & \\
        \hline
         \multirow{2}{4.4cm}{\textit{Simple\phantom{e}(D\phantom{-}Wiki+NewselaSL +simple+read)}}  &  \multirow{2}{8.5cm}{...Name are that the turtles go about nesting in a \textbf{strange} way... } \\
         & \\
         \hline
         \multirow{2}{4.4cm}{\textit{Simple\phantom{e}(D\phantom{-}Wiki+NewselaSL +simple+read+coh)):}}  &  \multirow{2}{8.5cm}{...Name are that the turtles go about nesting in a \textbf{peculiar} way... } \\
         & \\
         \hline
    \end{tabular}
    \caption{Example 3: Manual analysis for simplification models using D-Wikipedia and NewselaSL.}
    \label{tab:lexical_syntactic_example_3}
\end{table*}

\end{document}